\documentclass[10pt,twocolumn,letterpaper]{article}
\usepackage[pagenumbers]{cvpr} % To force page 
\usepackage{amssymb} % for arrow symbols
\usepackage{float} % Add this in the preamble
\usepackage{times}
\usepackage{latexsym}
\usepackage{microtype}
\usepackage{graphicx}
\usepackage{booktabs} % for professional tables
\usepackage{siunitx}
\usepackage{amsmath}
\usepackage{mathtools}
\usepackage{amsthm}
\usepackage{pdflscape,longtable,multirow}
\newcommand{\pos}[1]{\textcolor{ForestGreen}{#1}}
\newcommand{\nego}[1]{\textcolor{Orange}{#1}}

\newcommand{\bests}[1]{\textcolor{LimeGreen}{#1}}
\newcommand{\bads}[1]{\textcolor{red}{#1}}
\DeclareSIUnit\angstrom{\text {Å}}
\definecolor{cvprblue}{rgb}{0.21,0.49,0.74}
\usepackage[pagebackref,breaklinks,colorlinks,allcolors=cvprblue]{hyperref}
\def\paperID{7zraq6lnAH} % *** Enter the Paper ID here
\def\confName{CVPR}
\def\confYear{2025}

\title{Stress-Testing Multimodal Foundation Models for\\ Crystallographic Reasoning}
\author{%
Can Polat$^{1}$, Hasan Kurban$^{2}$\thanks{ Corresponding authors.}, Erchin Serpedin$^{1}$, and Mustafa Kurban$^{3,4*}$\\[1mm]
\footnotesize
$^{1}$Dept. of Electrical and Computer Engineering, Texas A\&M University, College Station, TX, USA\\
\footnotesize
$^{2}$College of Science and Engineering, Hamad Bin Khalifa University, Doha, Qatar\\
\footnotesize
$^{3}$Dept. of Electrical and Computer Engineering, Texas A\&M University at Qatar, Doha, Qatar\\
\footnotesize
$^{4}$Dept. of Prosthetics and Orthotics, Ankara University, Ankara, Turkey\\[2mm]
\footnotesize
\texttt{\{can.polat, eserpedin\}@tamu.edu, hkurban@hbku.edu.qa, kurbanm@ankara.edu.tr}
}

\usepackage{fancyhdr}

\fancypagestyle{accepted}{
  \fancyhf{}                             % clear all header and footer fields
}

\begin{document}
\maketitle
\begin{abstract}
Evaluating foundation models for crystallographic reasoning requires benchmarks that isolate generalization behavior while enforcing physical constraints. This work introduces a multiscale multicrystal dataset with two physically grounded evaluation protocols to stress-test multimodal generative models. The \textit{Spatial-Exclusion} benchmark withholds all supercells of a given radius from a diverse dataset, enabling controlled assessments of spatial interpolation and extrapolation. The \textit{Compositional-Exclusion} benchmark omits all samples of a specific chemical composition, probing generalization across stoichiometries. Nine vision--language foundation models are prompted with crystallographic images and textual context to generate structural annotations. Responses are evaluated via (i) relative errors in lattice parameters and density, (ii) a physics-consistency index penalizing volumetric violations, and (iii) a hallucination score capturing geometric outliers and invalid space-group predictions. These benchmarks establish a reproducible, physically informed framework for assessing generalization, consistency, and reliability in large-scale multimodal models. Dataset and code are available at \url{https://github.com/KurbanIntelligenceLab/StressTestingMMFMinCR}.
\end{abstract}

\section{Introduction}

Crystalline solids underpin a wide range of modern technologies. Their periodic atomic arrangements determine the band gaps of semiconductors, the ion-transport channels in battery electrodes, and the phonon spectra that govern thermal conductivity in microelectronics~\citep{wyckoff1963zno,bhadeshia2001geometry}. Even a single misassigned lattice parameter can cascade through simulation pipelines, distorting derived physical models and impeding materials discovery~\citep{levi,lubarda2003effective}. Structural resolution has traditionally relied on labor-intensive diffraction techniques or exhaustive structure enumeration followed by density functional theory (DFT) relaxation~\citep{kohn1965self}. Synthesis methods such as hydrothermal growth \citep{baruah2009hydrothermal}, chemical vapor deposition \citep{carlsson2010chemical}, and high-pressure processing \citep{bertucco2001high} further introduce domain-specific variability by accessing distinct thermodynamic regimes and defect topologies.

Recent progress in generative modeling, particularly autoregressive language models capable of emitting crystallographic information files~\citep{hall1991crystallographic}, enables rapid lattice generation with chemically plausible compositions. However, existing materials databases—such as AFLOW~\citep{aflow}, the Materials Project~\citep{jain2013commentary}, and OQMD~\citep{saal2013materials}—remain predominantly unimodal and typically lack expert-written, human-interpretable descriptions of crystal chemistry. This absence of multimodality impedes systematic evaluation of large vision–language models and language models in crystallographic reasoning. Current scientific multimodal benchmarks are limited in scale, visually simplistic, and textually sparse, constraining analysis of factual accuracy, hallucination patterns, and compliance with physical laws.

To overcome these limitations, a new multimodal dataset of crystalline materials is introduced, accompanied by two physically grounded benchmarking protocols. The \emph{spatial-exclusion} (SE) benchmark withholds supercells of a specific radius from the set \(\{R_k\}_{k=7}^{10}\), enabling controlled evaluation of spatial interpolation (interior radii) and extrapolation (boundary radii). In parallel, the \emph{compositional-exclusion} (CE) benchmark withholds all samples corresponding to a target chemical composition, assessing generalization across compositional space. State-of-the-art foundation models are evaluated under both benchmarks by generating structural annotations from crystallographic images and textual prompts. Model outputs are parsed into a structured \textsc{Material Properties} schema and assessed for geometric accuracy, consistency with physical constraints, and hallucination risk. These benchmarks provide a reproducible, domain-informed framework for measuring generalization and reliability in large-scale generative models, and contribute to emerging efforts to probe, refine, and safely deploy scientific knowledge at scale.

The remainder of the manuscript is structured as follows. Section \ref{sec:bg} surveys the theoretical foundations and related literature. Section \ref{sec:methods} details the methodological framework. Section \ref{sec:experiments} describes the dataset construction, evaluation metrics, and experimental procedures. Section \ref{sec:results} presents the empirical findings. Section \ref{sec:limits} discusses the study’s limitations, and Section \ref{sec:conc} concludes with final observations.  

\section{Background}
\label{sec:bg}
\subsection{Materials Simulation: From DFT to DFTB}

DFT offers first-principles access to total energies, atomic forces, and electronic densities of periodic solids, making it the gold standard for \textit{ab initio} crystal structure prediction and property evaluation~\citep{jensen2018numerical}. However, its cubic scaling with respect to basis-set size renders large-scale supercell studies computationally intractable in high-throughput campaigns involving thousands of candidate lattices~\citep{hourahine2007self}. To address this limitation, semi-empirical methods such as density functional tight binding (DFTB)~\citep{gaus2011dftb3} expand the Kohn–Sham energy around a reference density. The resulting Hamiltonians retain much of DFT's accuracy while reducing computational cost by two to three orders of magnitude. Recent advances—including self-consistent charge corrections and Slater–Koster parameterizations—have further extended DFTB’s applicability to heavy elements, excited-state phenomena, and non-adiabatic dynamics~\citep{papaconstantopoulos2003slater}.

\subsection{Machine Learning Models in Materials Science}

Machine learning models increasingly complement first-principles simulations in materials discovery pipelines. Neural architectures, particularly graph neural networks~\citep{zheng2018machine,rane2023transformers,liao2023equiformerv2,kurban2024enhancing}, operate directly on atomic graphs to predict material properties with high accuracy~\citep{schnet,dimenet,du2024densegnn,faenet}. Despite their predictive power, the deployment of such models in automated discovery frameworks remains constrained by data heterogeneity and computational bottlenecks. Recent multimodal approaches~\citep{gong2023multimodal,polat2024multimodal,li2025chemvlm} aim to unify structural, visual, and textual modalities to improve prediction fidelity and interpretability.

Multiple domain-specific benchmarks have emerged to support this integration. ScienceQA~\citep{lu2022learn} offers over 21,000 multimodal science questions grounded in school curricula, combining text, images, and expert annotations. MoleculeNet~\citep{wu2018moleculenet} standardizes molecular property prediction tasks across quantum chemistry, biophysics, and biomedical domains. LAB-Bench~\citep{laurent2024lab} includes more than 2,400 biology questions covering figure interpretation, protocol planning, and literature retrieval. TDCM25~\citep{polat2025tdcm25} spans approximately 100,000 TiO$_2$ structures across anatase, brookite, and rutile phases, integrating data from 0--1000~K for multitask benchmarking. ChemLit-QA~\citep{wellawatte2024chemlit} provides expert-validated QA pairs with hallucination and multi-hop reasoning annotations. Applications such as MatterChat \citep{tang2025matterchat}, xChemAgents \citep{polat2025xchemagents}, and  HoneyComb~\citep{zhang2024honeycomb} introduces an LLM-driven frameworks equipped with curated materials knowledge and dynamic tool integration for workflow orchestration. However, their capabilities are still limited \citep{miret2024llms}.

\section{Method}
\label{sec:methods}
\subsection{Crystal Generation}
\label{sec:crystal_gen}
The face‐centered cubic (FCC) lattice constant of gold is defined as  
\[
a_{\mathrm{Au}} \;=\; \SI{4.0782}{\angstrom},
\]
with space group Fm\(\overline{3}\)m~\citep{gold2002crc}.  

The corresponding primitive lattice vectors are:  
\begin{align*}
\mathbf{a}_1^{(\mathrm{Au})} &= \frac{a_{\mathrm{Au}}}{2}\,(0,\,1,\,1), \nonumber\\
\mathbf{a}_2^{(\mathrm{Au})} &= \frac{a_{\mathrm{Au}}}{2}\,(1,\,0,\,1), \label{eq:au_lattice_vectors}
\\
\mathbf{a}_3^{(\mathrm{Au})} &= \frac{a_{\mathrm{Au}}}{2}\,(1,\,1,\,0). \nonumber
\end{align*}
These define the $3\times3$ lattice matrix:  
\[
\mathbf{A}^{(\mathrm{Au})} 
\;=\; 
\bigl[\mathbf{a}_1^{(\mathrm{Au})}\;\mathbf{a}_2^{(\mathrm{Au})}\;\mathbf{a}_3^{(\mathrm{Au})}\bigr]
\;\in\;\mathbb{R}^{3\times 3}.
\]
The gold atom in the primitive cell is located at:  
\[
\mathbf{r}_0^{(\mathrm{Au})} \;=\; (0,\,0,\,0).
\]

To construct a spherical supercell of radius \(R\) (in nanometers), an integer multiplicity matrix is defined:

\begin{align*}
&\mathbf{S} 
= 
\begin{bmatrix}
s_{11} & s_{12} & s_{13} \\
s_{21} & s_{22} & s_{23} \\
s_{31} & s_{32} & s_{33}
\end{bmatrix}, \\ 
\quad &\mathbf{S}\in\mathbb{Z}_{>0}^{3\times3}, \quad \det\bigl(\mathbf{S}\bigr)\le 8,
\end{align*}
which ensures that the supercell volume is at most eight times that of the primitive cell.

Each lattice index is:

\[
\mathbf{n}
\;=\;
\begin{pmatrix}
n_1 \\
n_2 \\
n_3
\end{pmatrix}
\;\in\;\mathbb{Z}^3,
\]
and its Cartesian position is computed by:
\begin{equation}\label{eq:au_position_mapping}
\mathbf{r}' 
\;=\; 
\mathbf{A}^{(\mathrm{Au})}\,\mathbf{S}\,\mathbf{n}.
\end{equation}

The atomic positions are retained if they satisfy the distance criterion:
\begin{equation}\label{eq:au_distance_criterion}
\bigl\lVert \mathbf{r}' - \mathbf{r}_0^{(\mathrm{Au})} \bigr\rVert_2 \;\le\; R.
\end{equation}

For each \(R \in \{0.7,\,0.8,\,0.9,\,1.0\}\)~nanometer (nm), the smallest integer matrix \(\mathbf{S}\) with \(\det(\mathbf{S}) \le 8\) is selected to ensure the supercell encloses all atoms within distance \(R\).

The same process—computing lattice vectors \(\mathbf{A}^{(m)}\), selecting \(\mathbf{S}\), mapping index vectors via Equation~\eqref{eq:au_position_mapping}, and applying the retention criterion from Equation~\eqref{eq:au_distance_criterion}—is applied to nine additional materials: Ag, CH\(_3\)NH\(_3\)PbI\(_3\), Fe\(_2\)O\(_3\), MoS\(_2\), PbS, SnO\(_2\), SrTiO\(_3\), TiO\(_2\) (anatase), and ZnO. Unit-cell structures are obtained from experimental data while nanoparticles generated in-house, with full experimental data details listed in Appendix~\ref{appen:crystal}.

\subsection{Uniform Orientation Sampling and Rendering}
\label{subsec:rotations_rendering}

To obtain nine evenly spaced rotation axes, the Fibonacci-sphere method~\citep{stanley1975fibonacci} is applied with \(N=9\). Let \(\varphi = (1 + \sqrt{5})/2\). For \(k = 0,1,\dots,8\),
\begin{align*}
z_k &= 1 - \frac{2k + 1}{N}, \\
r_k &= \sqrt{1 - z_k^2},  \\
\phi_k &= 2\pi\,k\,\varphi^{-1} \pmod{2\pi}, \\
\hat{\mathbf{u}}_k &= \left(r_k \cos\phi_k,\, r_k \sin\phi_k,\, z_k\right) \in \mathbb{S}^2.
\end{align*}

Each axis \(\hat{\mathbf{u}}_k\) is associated with a fixed rotation angle \(\theta = \SI{30}{\degree}\). The corresponding rotation matrix \(R_k \in \mathrm{SO}(3)\) is given by Rodrigues’ formula:

\begin{align*}
R_k = I_3 \cos\theta 
    + (1 - \cos\theta)\,\hat{\mathbf{u}}_k\,\hat{\mathbf{u}}_k^{\!\top}  
    + \sin\theta\,\bigl[\hat{\mathbf{u}}_k\bigr]_\times.
\end{align*}

where
\[
[\hat{\mathbf{u}}_k]_\times =
\begin{pmatrix}
0 & -u_{k,3} & u_{k,2} \\
u_{k,3} & 0 & -u_{k,1} \\
-u_{k,2} & u_{k,1} & 0
\end{pmatrix}.
\]

Given \(N\) atomic positions \(\mathcal{P} = \{\mathbf{r}_i\}_{i=1}^N\), the center of mass is
\[
\mathbf{c} = \frac{1}{N} \sum_{i=1}^N \mathbf{r}_i.
\]
The rotated configuration is:
\[
\mathcal{P}^{(k)} = \left\{ R_k(\mathbf{r}_i - \mathbf{c}) + \mathbf{c} \;\middle|\; i = 1, \dots, N \right\}.
\]

Each rotated set \(\mathcal{P}^{(k)}\) is orthographically projected onto the \(xy\)-plane:
\[
(x_i^{(k)},\,y_i^{(k)},\,z_i^{(k)}) \mapsto (x_i^{(k)},\,y_i^{(k)}).
\]

Atoms are visualized as Gaussian-blurred disks with radius proportional to their covalent radius \(r_{\mathrm{cov}}(a)\) and colored using a CPK-like palette \(c(a)\). This pipeline produces nine consistent, rotation-augmented images per supercell, enabling model evaluation without stochastic variation.

\subsection{Expert Annotation Schema}

Each structure is paired with a structured textual record listing: atom count, supercell volume \(V = abc\), lattice parameters \((a,b,c)\), average nearest-neighbor distance \(\langle r_{\rm NN} \rangle\), bulk density \(\rho = m_{\rm tot} / V\), primitive-cell parameters \((a_0,b_0,c_0,\alpha,\beta,\gamma)\), space-group symbol, and a descriptive paragraph. This schema supports evaluation of geometric precision, physical-law consistency, hallucination rate, and rotation invariance.

\section{Experiments}
\label{sec:experiments}
\paragraph{Dataset.}

\begin{figure*}[ht]
    \centering
    \includegraphics[width=1\linewidth]{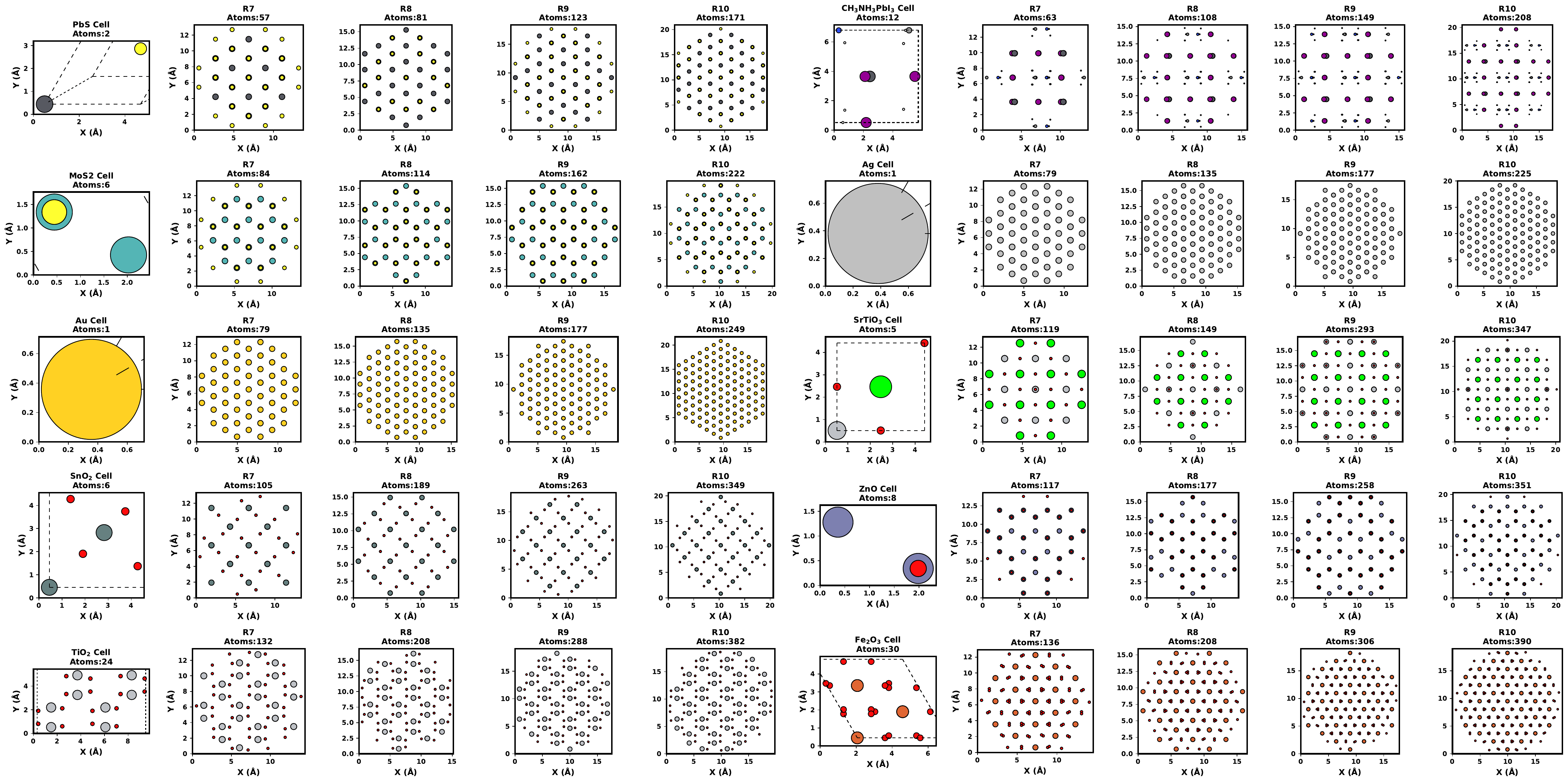}
    \caption{Gallery of atomic structures for each material. The first column shows the primitive unit cell for each material, while the subsequent columns display cluster structures with increasing radii ($R7$, $R8$, $R9$, $R10$). Each structure is visualized in a canonical orientation, with the number of atoms indicated in each panel. Materials are sorted by the atom count of their largest ($R10$) cluster.}
    \label{fig:materials}
\end{figure*}

The corpus comprises ten crystalline compounds of technological relevance—Ag, Au, CH$_3$NH$_3$PbI$_3$, Fe$_2$O$_3$, MoS$_2$, PbS, SnO$_2$, SrTiO$_3$, TiO$_2$, and ZnO—each relaxed using the SCC-DFTB formalism described in Section~\ref{sec:crystal_gen}. For each material, supercells of four target radii $R \in \{0.7, 0.8, 0.9, 1.0\}\,\mathrm{nm}$ (denoted as R7–R10) were generated, producing structures ranging from 57 to 390 atoms. These sizes bracket both experimentally reported and strained configurations. Each relaxed supercell was rendered in ten orientations: the native pose plus nine Fibonacci-sphere rotations to ensure quasi-uniform sampling of $\mathrm{SO}(3)$. The resulting dataset includes XYZ coordinates, $64 \times 64$-pixel JPEGs, and expert annotations compliant with the \textsc{Material Properties} schema. An overview is provided in Figure~\ref{fig:materials}.

\noindent
\paragraph{Evaluation Metrics.}\label{subsec:metrics}
\noindent
\textsc{Percent error} for each numerical property 
$p \in \{N_{\rm atoms},\,V_{\rm cell},\,a,\,b,\,c,\,\rho,\,a_{\rm p},\,b_{\rm p},\,c_{\rm p}\}$
is computed as:
\noindent
\begin{align*}
\Delta_p~[\%] =  100 \cdot \frac{|p^{\mathrm{gen}} - p^{\mathrm{ref}}|}{|p^{\mathrm{ref}}|}. 
\end{align*}
\noindent
\textsc{Space‐group match} is defined as:
\begin{equation*}
I_{\mathrm{SG}} = \mathbf{1}\bigl(\mathrm{SG}^{\mathrm{gen}} = \mathrm{SG}^{\mathrm{ref}}\bigr).
\end{equation*}
\noindent
Group statistics over $n$ examples are:
\begin{align*}
\mu_p &= \frac{1}{n}\sum_{i=1}^n \%\Delta_p^{(i)}, \\
\sigma_p &= \sqrt{\frac{1}{n-1} \sum_{i=1}^n \left(\%\Delta_p^{(i)} - \mu_p\right)^2}, \\
\mathrm{CI}_{95} &= \mu_p \pm 1.96 \cdot \frac{\sigma_p}{\sqrt{n}}.
\end{align*}
\noindent
\textsc{Prediction consistency (rotations)}  is computed by:
\begin{equation*}
C_{\mathrm{pred}} = 1 - \min\left(\frac{\sigma_r}{\mu_r},\,1\right),
\end{equation*}
where $\mu_r$ and $\sigma_r$ are the mean and standard deviation of a rotation-specific error set.\\
\noindent
\textsc{Physical‐law compliance} is evaluated for:
\[
p \in \left\{\rho,\frac{b}{a},\frac{c}{a},\left(\frac{b}{a}\right)_{\mathrm{prim}},\left(\frac{c}{a}\right)_{\mathrm{prim}}\right\},
\]
using:
\begin{align*}
\delta_p &= \frac{|p^{\mathrm{gen}} - p^{\mathrm{ref}}|}{p^{\mathrm{ref}}}, \\
s_p &=
\begin{cases}
1.0 & \delta_p \le 0.10, \\
0.5 & 0.10 < \delta_p \le 0.25, \\
0.0 & \delta_p > 0.25 \text{ or on error}.
\end{cases}
\end{align*}
Aggregate score:
\begin{equation*}
S_{\mathrm{phys}} =
\begin{cases}
\frac{1}{N} \sum_p s_p & N > 0, \\
0.0 & N = 0 \text{ or missing}.
\end{cases}
\end{equation*}
\noindent
\textsc{Hallucination score} is defined for all the percent error properties $p$.
Let $g = p^{\mathrm{gen}}$ and $r = p^{\mathrm{ref}}$, then:
\begin{equation*}
h_p =
\begin{cases}
1.0 & g \le 0 \quad \text{(non-physical)}, \\
1.0 & \frac{|g - r|}{|r|} > 0.25, \\
0.5 & 0.10 < \frac{|g - r|}{|r|} \le 0.25, \\
0.0 & \frac{|g - r|}{|r|} \le 0.10.
\end{cases}
\end{equation*}
Let $M$ be the number of valid checks:
\begin{equation*}
S_{\mathrm{hall}} =
\begin{cases}
\frac{1}{M} \sum_p h_p & M > 0, \\
0.0 & M = 0, \\
1.0 & \text{if input is \texttt{None}}.
\end{cases}
\end{equation*}
Additional metric definitions are provided in Appendix~\ref{appen:extendMetric}.

\paragraph{Spatial-Exclusion Protocol.}
SE protocol measures extrapolation across length scales. For each material $m_i$ with radius set $\mathcal{R}_{m_i}$, one radius $R_* \in \mathcal{R}_{m_i}$ is held out. The model context includes:
\[
\bigl|\mathcal{R}_{m_i}\setminus\{R_*\}\bigr|\times 5
\]
examples (5 rotations for each of the remaining radii). Each test instance uses only the Cartesian coordinates of $(m_i, R_*, k)$, and the model must generate predictions without seeing any data at $R_*$. The overall SE error is:\[
E_{\mathrm{SE}}
= \frac{1}{%
  |\mathcal{M}|\;\sum_{i}|\mathcal{R}_{m_i}|\;\times 5%
}
\]
\[
\qquad\quad
\times \sum_{i}\;\sum_{R_*\in\mathcal{R}_{m_i}}\;\sum_{k=0}^{4}
\ell\bigl(\hat y_{i,R_*,k},\,y_{i,R_*,k}\bigr),
\]
where $\ell$ is the percent error loss.

\paragraph{Compositional-Exclusion Protocol. }
CE protocol assesses generalization across compositions. For each material $m_i$, all of its data are excluded from the context. The context size becomes:
\[
\Bigl(\sum_{m_j\neq m_i}|\mathcal{R}_{m_j}|\Bigr)\times 5
\]
At test time, only the Cartesian coordinates of $(m_i, R_*, k)$ are given. The transfer error is:
\[
E_{\mathrm{CE}}
= \frac{1}{%
  |\mathcal{M}|\;\sum_{i}|\mathcal{R}_{m_i}|\;\times 5%
}
\]
\[
\qquad\quad
\times \sum_{i}\;\sum_{R_*\in\mathcal{R}_{m_i}}\;\sum_{k=0}^{4}
\ell\bigl(\tilde y_{i,R_*,k},\,y_{i,R_*,k}\bigr),
\]
which captures model performance when required to infer from disjoint compositions. Comparing $E_{\mathrm{CE}}$ and $E_{\mathrm{SE}}$ helps isolate failure modes in spatial vs. chemical generalization.

\begin{table*}[ht]
  \centering
  %––– Part (a) –––
  \small
  \noindent{(a) Spatial-Exclusion (SE)} 
  \resizebox{\textwidth}{!}{%
  \begin{tabular}{lcccccccccccccccccccccccc}
    \toprule
    Material & \multicolumn{6}{c}{R7} & \multicolumn{6}{c}{R8}
             & \multicolumn{6}{c}{R9} & \multicolumn{6}{c}{R10} \\
     & $\%\Delta N_A$ & $\%\Delta V$ & $\%\Delta a$ & $\%\Delta b$ & $\%\Delta c$ & $\%\Delta \rho$
     & $\%\Delta N_A$ & $\%\Delta V$ & $\%\Delta a$ & $\%\Delta b$ & $\%\Delta c$ & $\%\Delta \rho$
     & $\%\Delta N_A$ & $\%\Delta V$ & $\%\Delta a$ & $\%\Delta b$ & $\%\Delta c$ & $\%\Delta \rho$
     & $\%\Delta N_A$ & $\%\Delta V$ & $\%\Delta a$ & $\%\Delta b$ & $\%\Delta c$ & $\%\Delta \rho$ \\
    \cmidrule(lr){2-7}
    \cmidrule(lr){8-13}
    \cmidrule(lr){14-19}
    \cmidrule(lr){20-25}
        Ag & $26.53$ & $46.74$ & $9.21$ & $13.00$ & $21.32$ & $14.00$ & $10.21$ & $14.59$ & $5.11$ & $5.98$ & $8.12$ & $13.63$ & $7.31$ & $15.00$ & $7.88$ & $8.52$ & $\bads{10.05}$ & $7.96$ & $7.48$ & $9.64$ & $5.65$ & $5.07$ & $9.81$ & $8.36$ \\
        Au & $28.21$ & $49.44$ & $10.18$ & $13.44$ & $22.48$ & $15.47$ & $11.26$ & $14.20$ & $5.43$ & $6.55$ & $6.42$ & $11.38$ & $9.19$ & $12.26$ & $6.26$ & $7.51$ & $9.37$ & $8.58$ & $15.40$ & $\bads{583.53}$ & $\bads{40.45}$ & $\bads{40.73}$ & $\bads{44.04}$ & $17.61$ \\
        CH$_3$NH$_3$PbI$_3$ & $\bads{47.34}$ & $44.81$ & $\bads{16.10}$ & $10.85$ & $12.31$ & $34.38$ & $16.83$ & $20.28$ & $\bads{7.45}$ & $\bads{7.82}$ & $7.08$ & $20.85$ & $17.58$ & $\bads{27.85}$ & $8.52$ & $\bads{9.34}$ & $8.51$ & $22.72$ & $13.45$ & $19.11$ & $8.94$ & $8.44$ & $9.11$ & $\bads{128.49}$ \\
        Fe$_2$O$_3$ & $26.21$ & $31.41$ & $8.92$ & $10.46$ & $12.99$ & $13.37$ & $13.42$ & $20.18$ & $6.80$ & $\bests{4.00}$ & $7.31$ & $11.34$ & $12.23$ & $15.81$ & $5.84$ & $6.60$ & $5.18$ & $10.46$ & $11.92$ & $12.86$ & $4.45$ & $5.23$ & $4.97$ & $6.93$ \\
        MoS$_2$ & $15.48$ & $27.46$ & $9.21$ & $9.17$ & $\bads{22.55}$ & $10.29$ & $16.80$ & $14.31$ & $5.84$ & $5.53$ & $\bads{11.78}$ & $16.10$ & $9.59$ & $17.76$ & $7.18$ & $7.67$ & $7.22$ & $9.63$ & $\bests{5.69}$ & $19.28$ & $5.63$ & $7.79$ & $10.97$ & $19.16$ \\
        PbS & $17.54$ & $29.71$ & $9.07$ & $10.78$ & $11.75$ & $\bads{39.28}$ & $18.66$ & $\bads{23.90}$ & $6.16$ & $7.38$ & $11.53$ & $19.91$ & $12.90$ & $22.60$ & $\bads{9.69}$ & $8.78$ & $9.62$ & $\bads{28.27}$ & $12.27$ & $14.45$ & $7.25$ & $5.99$ & $8.00$ & $13.85$ \\
        SnO$_2$ & $29.48$ & $19.31$ & $8.25$ & $7.02$ & $\bests{10.03}$ & $26.84$ & $9.78$ & $18.99$ & $\bests{4.31}$ & $4.04$ & $9.33$ & $7.56$ & $8.24$ & $12.57$ & $5.32$ & $\bests{5.25}$ & $6.90$ & $7.21$ & $6.80$ & $10.86$ & $\bests{4.42}$ & $4.15$ & $8.57$ & $8.34$ \\
        SrTiO$_3$ & $30.59$ & $\bads{55.64}$ & $15.59$ & $\bads{16.05}$ & $15.82$ & $17.99$ & $\bads{37.42}$ & $20.10$ & $7.12$ & $7.77$ & $8.25$ & $\bads{38.31}$ & $\bads{20.30}$ & $22.56$ & $7.87$ & $7.19$ & $7.56$ & $17.26$ & $\bads{21.58}$ & $21.49$ & $6.84$ & $6.80$ & $7.36$ & $16.79$ \\
        TiO$_2$ & $23.54$ & $22.99$ & $\bests{6.71}$ & $\bests{6.27}$ & $12.77$ & $\bests{6.42}$ & $\bests{8.08}$ & $\bests{9.54}$ & $4.48$ & $4.09$ & $\bests{4.35}$ & $\bests{5.09}$ & $6.39$ & $\bests{9.32}$ & $\bests{4.88}$ & $5.92$ & $4.75$ & $\bests{6.35}$ & $5.76$ & $\bests{6.95}$ & $4.84$ & $\bests{4.12}$ & $\bests{3.44}$ & $\bests{5.48}$ \\
        ZnO & $\bests{13.11}$ & $\bests{16.66}$ & $10.47$ & $9.28$ & $12.22$ & $21.97$ & $12.74$ & $12.42$ & $4.96$ & $5.23$ & $6.98$ & $11.04$ & $\bests{5.66}$ & $9.63$ & $5.05$ & $6.01$ & $\bests{4.59}$ & $8.50$ & $8.91$ & $19.57$ & $6.21$ & $7.49$ & $8.89$ & $20.74$ \\
    \bottomrule
  \end{tabular}%
  }
  
  \bigskip
  
  %––– Part (b) –––
  \small
  \noindent{(b) Compositional-Exclusion (CE)} 
  \resizebox{\textwidth}{!}{%
  \begin{tabular}{lcccccccccccccccccccccccccccc}
    \toprule
    Material & \multicolumn{7}{c}{R7} & \multicolumn{7}{c}{R8}
             & \multicolumn{7}{c}{R9} & \multicolumn{7}{c}{R10} \\
     & $\%\Delta N_A$ & $\%\Delta a_p$ & $\%\Delta b_p$ & $\%\Delta c_p$
     & $|\Delta \alpha_p|$ & $|\Delta \beta_p|$ & $|\Delta \gamma_p|$
     & $\%\Delta N_A$ & $\%\Delta a_p$ & $\%\Delta b_p$ & $\%\Delta c_p$
     & $|\Delta \alpha_p|$ & $|\Delta \beta_p|$ & $|\Delta \gamma_p|$
     & $\%\Delta N_A$ & $\%\Delta a_p$ & $\%\Delta b_p$ & $\%\Delta c_p$
     & $|\Delta \alpha_p|$ & $|\Delta \beta_p|$ & $|\Delta \gamma_p|$
     & $\%\Delta N_A$ & $\%\Delta a_p$ & $\%\Delta b_p$ & $\%\Delta c_p$
     & $|\Delta \alpha_p|$ & $|\Delta \beta_p|$ & $|\Delta \gamma_p|$ \\
    \cmidrule(lr){2-8}
    \cmidrule(lr){9-15}
    \cmidrule(lr){16-22}
    \cmidrule(lr){23-29}
        Ag & $6.39$ & $10.49$ & $10.49$ & $10.49$ & $7.50$ & $7.50$ & $7.50$ & $\bests{3.39}$ & $10.48$ & $10.48$ & $10.48$ & $7.50$ & $7.50$ & $7.50$ & $4.28$ & $10.47$ & $10.47$ & $10.47$ & $7.50$ & $7.50$ & $7.50$ & $14.09$ & $13.76$ & $13.76$ & $13.76$ & $6.75$ & $6.75$ & $6.75$ \\
        Au & $\bests{3.58}$ & $17.79$ & $17.79$ & $17.79$ & $6.00$ & $6.00$ & $6.00$ & $4.89$ & $15.63$ & $15.63$ & $15.63$ & $4.50$ & $4.50$ & $4.50$ & $\bests{3.95}$ & $16.76$ & $16.76$ & $16.76$ & $4.50$ & $4.50$ & $4.50$ & $\bests{12.64}$ & $16.65$ & $16.65$ & $16.65$ & $4.50$ & $4.50$ & $4.50$ \\
        CH$_3$NH$_3$PbI$_3$ & $\bads{32.14}$ & $10.48$ & $10.46$ & $23.36$ & $1.50$ & $1.86$ & $3.75$ & $37.87$ & $5.04$ & $9.27$ & $9.52$ & $4.68$ & $5.04$ & $4.68$ & $46.59$ & $11.51$ & $16.15$ & $21.65$ & $3.07$ & $3.43$ & $3.07$ & $\bads{47.69}$ & $12.27$ & $12.39$ & $10.45$ & $1.55$ & $1.91$ & $6.80$ \\
        Fe$_2$O$_3$ & $18.42$ & $2.95$ & $1.10$ & $8.65$ & $1.50$ & $1.50$ & $5.25$ & $27.82$ & $\bests{2.79}$ & $2.77$ & $9.70$ & $0.75$ & $0.78$ & $5.25$ & $26.00$ & $3.46$ & $3.46$ & $11.38$ & $1.50$ & $1.50$ & $6.00$ & $19.60$ & $5.01$ & $3.36$ & $10.34$ & $1.74$ & $1.74$ & $6.24$ \\
        MoS$_2$ & $13.57$ & $\bests{0.04}$ & $\bests{0.04}$ & $\bests{0.02}$ & $\bests{0.00}$ & $\bests{0.00}$ & $\bests{0.00}$ & $23.42$ & $7.45$ & $7.42$ & $3.72$ & $\bests{0.00}$ & $0.03$ & $2.25$ & $18.61$ & $\bests{0.01}$ & $\bests{0.01}$ & $0.02$ & $\bests{0.00}$ & $\bests{0.00}$ & $\bests{0.00}$ & $26.55$ & $0.04$ & $\bests{0.04}$ & $0.03$ & $\bests{0.00}$ & $\bests{0.00}$ & $\bests{0.00}$ \\
        PbS & $30.95$ & $40.57$ & $\bads{40.55}$ & $40.57$ & $\bads{22.31}$ & $\bads{22.33}$ & $\bads{22.31}$ & $\bads{56.05}$ & $40.13$ & $\bads{40.13}$ & $39.16$ & $\bads{24.00}$ & $\bads{24.00}$ & $\bads{24.00}$ & $\bads{58.68}$ & $41.16$ & $\bads{41.16}$ & $41.16$ & $\bads{24.75}$ & $\bads{24.75}$ & $\bads{24.75}$ & $44.59$ & $40.11$ & $\bads{40.11}$ & $40.11$ & $\bads{24.75}$ & $\bads{24.75}$ & $\bads{24.75}$ \\
        SnO$_2$ & $19.17$ & $4.71$ & $0.78$ & $3.08$ & $\bests{0.00}$ & $\bests{0.00}$ & $\bests{0.00}$ & $19.79$ & $5.67$ & $\bests{1.73}$ & $13.80$ & $\bests{0.00}$ & $\bests{0.01}$ & $0.75$ & $31.33$ & $2.42$ & $0.40$ & $1.59$ & $\bests{0.00}$ & $\bests{0.00}$ & $\bests{0.00}$ & $16.14$ & $2.36$ & $0.39$ & $1.55$ & $\bests{0.00}$ & $\bests{0.00}$ & $0.75$ \\
        SrTiO$_3$ & $27.72$ & $8.54$ & $4.44$ & $5.79$ & $0.43$ & $0.43$ & $0.43$ & $22.48$ & $11.82$ & $11.74$ & $13.02$ & $\bests{0.00}$ & $0.06$ & $\bests{0.00}$ & $27.52$ & $9.53$ & $3.81$ & $7.68$ & $\bests{0.00}$ & $\bests{0.00}$ & $0.60$ & $19.49$ & $1.53$ & $1.52$ & $1.54$ & $\bests{0.00}$ & $0.01$ & $\bests{0.00}$ \\
        TiO$_2$ & $21.02$ & $\bads{51.78}$ & $18.81$ & $\bads{55.18}$ & $\bests{0.00}$ & $\bests{0.00}$ & $1.50$ & $19.81$ & $\bads{51.80}$ & $19.62$ & $\bads{47.97}$ & $\bests{0.00}$ & $\bests{0.01}$ & $2.25$ & $32.06$ & $\bads{51.96}$ & $19.12$ & $\bads{53.58}$ & $\bests{0.00}$ & $\bests{0.00}$ & $1.50$ & $20.75$ & $\bads{53.15}$ & $21.12$ & $\bads{49.80}$ & $\bests{0.00}$ & $\bests{0.00}$ & $3.75$ \\
        ZnO & $24.53$ & $1.16$ & $2.66$ & $1.03$ & $\bests{0.00}$ & $\bests{0.00}$ & $1.50$ & $23.26$ & $5.86$ & $7.34$ & $\bests{2.07}$ & $\bests{0.00}$ & $0.02$ & $2.25$ & $31.50$ & $\bests{0.01}$ & $\bests{0.01}$ & $\bests{0.01}$ & $\bests{0.00}$ & $\bests{0.00}$ & $\bests{0.00}$ & $24.59$ & $\bests{0.01}$ & $3.02$ & $\bests{0.01}$ & $\bests{0.00}$ & $\bests{0.00}$ & $0.75$ \\
    \bottomrule
  \end{tabular}%
  }
    \caption{Mean percent errors (\%\(\Delta\)) for (a) the spatial-extension (SE) protocol—evaluating extrapolation to unseen supercell radii—and (b) the compositional-exclusion (CE) protocol—evaluating cross‐material transfer. Part (a) reports errors on atom count \(N_A\), cell volume \(V\), lattice parameters \(a\), \(b\), \(c\), and density \(\rho\); part (b) reports errors on \(N_A\), primitive cell edges \(a_p\), \(b_p\), \(c_p\), and absolute angular deviations \(|\Delta\alpha_p|\), \(|\Delta\beta_p|\), \(|\Delta\gamma_p|\). Results are shown for each material and radius value (R7–R10), averaged over five random rotations per configuration and across all models. Lower values indicate better agreement with reference structures. These complementary metrics illustrate the model’s capacity to capture atomic‐scale patterns across variations in supercell size and material composition. Contrasting SE and CE errors highlights whether performance limitations stem from radius extrapolation or cross‐material generalization. Colours indicate predictive difficulty: \bests{green} marks the material with the lowest prediction error (easiest to predict), while \bads{red} marks the material with the highest prediction error (hardest to predict).}
  \label{tab:combined_percent_errors}
\end{table*}

\section{Results}
\label{sec:results}
\paragraph{SE Evaluation.}
In SE evaluation, each language model was assigned the task of predicting a held-out radius value (\(R_7\)–\(R_{10}\)) for a given crystalline material, and its outputs for atom count (\(N_A\)), cell volume (\(V\)), lattice constants (\(a\), \(b\), \(c\)), and density (\(\rho\)) were compared against reference structures. Percent errors (\(\%\Delta\)) were averaged across all models and five random 3D orientations per configuration. As shown in Table~\ref{tab:combined_percent_errors}(a), the resulting error rates remain consistently high, particularly for key physical properties—exceeding thresholds that render predictions scientifically unreliable.

These discrepancies reveal a fundamental limitation: the models fail to internalize core geometric and physical constraints that govern crystal structures. The inability to extrapolate structural properties across radii highlights the need for architectural enhancements, including explicit domain constraints, physical priors, and robust error-correction strategies to prevent hallucinated outputs and enforce consistency in atomic-scale reasoning.

\paragraph{CE Evaluation.}
In the CE evaluation, each language model received structural data from nine materials at a fixed radius \(R\) and was tasked with predicting \(N_A\), primitive cell lengths (\(a_p\), \(b_p\), \(c_p\)), and angles (\(\alpha_p\), \(\beta_p\), \(\gamma_p\)) for a held-out material. To ensure robustness, predictions were averaged over five random 3D orientations and multiple model variants. As reported in Table~\ref{tab:combined_percent_errors}(b), percent errors in cell lengths frequently exceed 15\%, and atom count errors surpass 30\% for complex compounds at smaller radii—suggesting a failure to generalize geometric patterns across novel chemistries.

Additionally, absolute deviations in primitive angles often exceed 5° and reach beyond 20° in certain cases, reflecting substantial geometric inconsistencies and a tendency to hallucinate physical details. These results reinforce that purely data-driven training is insufficient for capturing atomic-scale regularities. Embedding explicit domain constraints, structured knowledge priors, and uncertainty-aware mechanisms is essential for enforcing physical plausibility and mitigating hallucination in generative crystallography.
\begin{table}
\centering
\small
\setlength{\tabcolsep}{3pt}
\resizebox{\columnwidth}{!}{%
\begin{tabular}{lcccccc}
\toprule
Model & SE & CE & $T \times 10^3$ & $G_{\max} \times 10$ & $t_{SE}$& $t_{CE}$\\
\midrule
Claude Opus 4 (Anthropic)      & 0.06 & 0.91   & \textbf{2.17}    & \underline{3.04}  &12.86 & 13.91\\
Claude Sonnet 4 (Anthropic)    & \textbf{0.04} & \underline{0.68}   & 3.93    & \underline{3.04}  &6.43  & 8.23\\
DeepSeek‐Chat (DeepSeek)       & 0.09 & 1.79   & 14.16   & 6.47  &24.97 & 13.71\\
GPT‐4.1 Mini (OpenAI)          & 0.18 & \textbf{0.53}   & \underline{2.63}    & 6.00  &8.08  & 7.26\\
Gemini 2.5 Flash (Google)      & \underline{0.05} & 1.32   & 21.38   & \underline{3.04}  &\textbf{3.06}  & \textbf{5.00}\\
Grok 2 (X.ai)                  & 0.07 & 2.34   & 15.55   & \underline{3.04}  &6.37  & 8.99\\
Grok 2 Vision (X.ai)           & 0.06 & 2.02   & 22.54   & 6.47  &7.32  & 9.50\\
Llama-4 Maverick (Meta)        & 0.09 & 0.89   & 3.70    & \textbf{3.00}  &\underline{4.33}  & \underline{6.72}\\
Mistral Medium 3 (Mistral AI)  & \underline{0.05} & 0.92   & 11.24   & \textbf{3.00}  &14.78 & 15.45\\
\bottomrule
\end{tabular}%
}
\caption{Transfer degradation analysis with mean percent errors (\%\,$\Delta$) for the SE and CE splits. $T=\text{CE}/\text{SE}$; $G_{\max}$ is the largest absolute error observed in any single prediction. $t_{SE}$ and $t_{CE}$ represents the each models latency in seconds for SE and CE task, respectively. \textbf{Bold} indicates the top-performing model, while \underline{underlining} denotes the runner-up.}
\label{tab:transfer_scores}
\end{table}

\paragraph{Knowledge Transfer.}
CE evaluation reveals that current multimodal LLMs rely heavily on memorized numeric templates rather than internalized crystallographic principles. In the control setting (SE), all eight models achieve low mean percent errors (\(0.04 \le \text{SE} \le 0.18\)). However, when evaluated on withheld compounds, performance collapses: the average error increases by several orders of magnitude, and the transfer ratio \(T = \text{CE}/\text{SE}\) surges from \(2.2 \times 10^{3}\) to \(2.3 \times 10^{4}\), with one model diverging entirely (\(T = \infty\)).

A consistent failure pattern emerges across systems: six models record their largest relative error on the primitive-cell \(b\)-axis (\(\%\Delta b_p\)), while the remainder fail on \(\%\Delta a_p\). PbS is the most challenging composition, ranked worst by all models except one, which instead fails on Fe\textsubscript{2}O\textsubscript{3}. The rock-salt symmetry of PbS demands reconciliation between cubic crystal geometry and its serialized representation; instead, most models generate inconsistent or arbitrary lattice parameters. These findings underscore that in-distribution performance does not imply genuine crystallographic reasoning. Even modest compositional perturbations destabilize the geometric priors learned by large-scale vision–language models, revealing a brittle foundation for generalization.

\begin{table*}[ht]
\centering
  %––– Part (a) –––
  \small
  \noindent{(a) SE $\Rightarrow$ CE } 
  \resizebox{\textwidth}{!}{%
  \begin{tabular}{l*{14}{c}}
    \toprule
    & $N_\mathrm{atoms}\!\leftrightarrow V$ & $V\!\leftrightarrow \bar\varepsilon$ & $V\!\leftrightarrow \rho$ & $\gamma_p\!\leftrightarrow \bar\varepsilon$ & $a\!\leftrightarrow \bar\varepsilon$ & $a\!\leftrightarrow \rho$ & $a\!\leftrightarrow b$ & $a_p\!\leftrightarrow b_p$ & $a_p\!\leftrightarrow c_p$ & $b\!\leftrightarrow \bar\varepsilon$ & $b\!\leftrightarrow \rho$ & $b_p\!\leftrightarrow c_p$ & $c\!\leftrightarrow \bar\varepsilon$ & $c\!\leftrightarrow \rho$ \\
    \midrule
    $\epsilon_{\text{SE}}$ & \pos{+0.28} & \pos{+0.81} & \pos{+0.34} & \nego{-0.00} & \pos{+0.52} & \pos{+0.13} & \pos{+0.32} & \pos{+0.09} & \pos{+0.09} & \pos{+0.50} & \pos{+0.17} & \pos{+0.09} & \pos{+0.57} & \pos{+0.21} \\
    $\epsilon_{\text{CE}}$ & \pos{+0.02} & \pos{+0.17} & \nego{-0.06} & \pos{+0.22} & \pos{+0.09} & \nego{-0.10} & \nego{-0.00} & \pos{+0.69} & \pos{+0.44} & \pos{+0.07} & \nego{-0.05} & \pos{+0.44} & \pos{+0.10} & \nego{-0.14} \\
    $\Delta$ & \nego{-0.27} & \nego{-0.64} & \nego{-0.40} & \pos{+0.22} & \nego{-0.44} & \nego{-0.22} & \nego{-0.32} & \pos{+0.59} & \pos{+0.35} & \nego{-0.43} & \nego{-0.22} & \pos{+0.35} & \nego{-0.47} & \nego{-0.35} \\
    \bottomrule
  \end{tabular}
}
  
  \bigskip
  
  %––– Part (b) –––
  \small
  \noindent{(b) CE $\Rightarrow$ SE} 
  \resizebox{\textwidth}{!}{%
  \begin{tabular}{l*{14}{c}}
    \toprule
    & $N_\mathrm{atoms}\!\leftrightarrow V$ & $V\!\leftrightarrow \bar\varepsilon$ & $V\!\leftrightarrow \rho$ & $\gamma_p\!\leftrightarrow \bar\varepsilon$ & $a\!\leftrightarrow \bar\varepsilon$ & $a\!\leftrightarrow \rho$ & $a\!\leftrightarrow b$ & $a_p\!\leftrightarrow b_p$ & $a_p\!\leftrightarrow c_p$ & $b\!\leftrightarrow \bar\varepsilon$ & $b\!\leftrightarrow \rho$ & $b_p\!\leftrightarrow c_p$ & $c\!\leftrightarrow \bar\varepsilon$ & $c\!\leftrightarrow \rho$ \\
    \midrule
    $\epsilon_{\text{CE}}$ & \pos{+0.02} & \pos{+0.17} & \nego{-0.06} & \pos{+0.22} & \pos{+0.09} & \nego{-0.10} & \nego{-0.00} & \pos{+0.69} & \pos{+0.44} & \pos{+0.07} & \nego{-0.05} & \pos{+0.44} & \pos{+0.10} & \nego{-0.14} \\
    $\epsilon_{\text{SE}}$ & \pos{+0.28} & \pos{+0.81} & \pos{+0.34} & \nego{-0.00} & \pos{+0.52} & \pos{+0.13} & \pos{+0.32} & \pos{+0.09} & \pos{+0.09} & \pos{+0.50} & \pos{+0.17} & \pos{+0.09} & \pos{+0.57} & \pos{+0.21} \\
    $\Delta$ & \pos{+0.27} & \pos{+0.64} & \pos{+0.40} & \nego{-0.22} & \pos{+0.44} & \pos{+0.22} & \pos{+0.32} & \nego{-0.59} & \nego{-0.35} & \pos{+0.43} & \pos{+0.22} & \nego{-0.35} & \pos{+0.47} & \pos{+0.35} \\
    \bottomrule
  \end{tabular}
  }
\caption{Largest shifts in \emph{error--error} correlation coefficients when transferring between SE and CE annotation protocols. Each sub-table displays the top 14 property pairs (ordered alphabetically) exhibiting the largest absolute changes in pairwise correlation, averaged over all models, materials, and $R7$–$R10$. Panel (a) shows the shift from SE to CE ($\Delta = \rho_{\mathrm{CE}} - \rho_{\mathrm{SE}}$), while panel (b) shows the reverse (CE to SE, $\Delta = \rho_{\mathrm{SE}} - \rho_{\mathrm{CE}}$). For each property pair, the table reports the correlation coefficients under each protocol and their difference $\Delta$. Cells are color‐coded: \pos{green} for positive $\Delta$ (stronger coupling under the target protocol) and \nego{red} for negative $\Delta$ (weaker coupling), highlighting which structural or physical property relationships are most sensitive to the choice of annotation protocol.}
\label{tab:corr_shift_alphabetical_both}
\end{table*}

\begin{table}[t]
\centering
\resizebox{\columnwidth}{!}{%
  \begin{tabular}{lcc}
    \toprule
    {Material} & {Physical Law Compliance} & {Hallucination Score} \\    \midrule
Ag & \underline{0.82} $\pm$ 0.03 & 0.21 $\pm$ 0.04 \\
Au & \textbf{0.84} $\pm$ 0.03 & 0.24 $\pm$ 0.02 \\
CH$_3$NH$_3$PbI$_3$ & 0.72 $\pm$ 0.03 & 0.42 $\pm$ 0.05 \\
Fe$_2$O$_3$ & 0.74 $\pm$ 0.03 & 0.23 $\pm$ 0.02 \\
MoS$_2$ & 0.78 $\pm$ 0.03 & \textbf{0.18} $\pm$ 0.01 \\
PbS & 0.77 $\pm$ 0.03 & 0.53 $\pm$ 0.02 \\
SnO$_2$ & 0.74 $\pm$ 0.03 & 0.24 $\pm$ 0.04 \\
SrTiO$_3$ & 0.77 $\pm$ 0.02 & 0.28 $\pm$ 0.03 \\
TiO$_2$ & 0.46 $\pm$ 0.02 & 0.43 $\pm$ 0.03 \\
ZnO & 0.77 $\pm$ 0.02 & \underline{0.21} $\pm$ 0.02 \\
    \bottomrule
  \end{tabular}%
}
\caption{Mean ± std physical‐law compliance and hallucination scores for each material, averaged over all models and five runs per material–radius under both SE and CE protocols. Physical‐law compliance measures adherence to fundamental structural constraints (e.g., density and lattice‐parameter ratios), while the hallucination score quantifies the frequency of non‐physical or highly erroneous predictions across a set of key properties. \textbf{Boldface} denotes the material with the highest prediction accuracy, while \underline{underlining} denotes the material with the second highest accuracy.}
\label{tab:hallu}
\end{table}

\paragraph{Correlation Shift.}
Table~\ref{tab:corr_shift_alphabetical_both} reports the average error–error correlation coefficients for fourteen property pairs under the SE and CE protocols, along with their differences. Notably, the transition from SE to CE increases the correlation between projected lattice constants \(a_p\) and \(b_p\) by \(0.59\), suggesting that prediction errors for these geometric features become more aligned when the model is exposed to entirely novel compositions. In contrast, the correlation between volume \(V\) and average formation energy \(\bar\varepsilon\) drops by \(-0.64\), indicating a breakdown in the learned volume–energy coupling under compositional generalization.

These shifts reverse when comparing CE to SE, confirming that the observed effects stem from the validation regime rather than intrinsic data asymmetries. This bidirectional sensitivity highlights a critical weakness: current foundation models preserve certain geometric relationships under run-wise exclusion but fail to maintain deeper physical dependencies—such as energetic coherence—when facing unfamiliar chemistries. The instability of error correlations under different evaluation settings undermines the robustness of model generalization and emphasizes the need for embedding invariant physical priors into model architecture and training.

\paragraph{Compliance and Hallucination.}
The models consistently struggle to enforce fundamental physical constraints and frequently fabricate ungrounded details, as quantified in Table~\ref{tab:hallu}. Physical-law compliance scores fall below acceptable thresholds for most materials, with particularly poor performance on TiO$_2$, where nearly half the predictions violate basic geometric or density-based relationships. Concurrently, hallucination scores indicate that a significant fraction of predicted properties—often over 40\%—deviate substantially from reference values or represent nonphysical outputs. The co-occurrence of constraint violations and fictitious property generation highlights systemic limitations in current architectures. These results reinforce the need for models that integrate structural priors, conservation rules, and uncertainty-aware mechanisms to produce physically plausible and trustworthy predictions at the atomic scale.

\paragraph{Model Latency.} Table~\ref{tab:transfer_scores} presents the average inference latencies per sample across the SE and CE protocols. Gemini 2.5 Flash exhibits the lowest latency, requiring only \SI{3.06}{\second} under SE and \SI{5.00}{\second} under CE, making it well-suited for time-sensitive applications such as high-throughput materials screening. Llama-4 Maverick and GPT-4.1 Mini follow in the next performance tier with moderate latency (\SIrange{4}{8}{\second}), while most other models cluster between \SIrange{6}{15}{\second}. DeepSeek-Chat is the slowest model in the SE evaluation (\SI{25}{\second}), and Mistral Medium 3 exhibits the highest latency in CE (\SI{15.5}{\second}). These trends broadly correlate with model size and architecture, where larger context windows and multimodal inputs tend to incur higher computational overhead. Although latency is not the primary evaluation criterion in this study, the results offer practical insights for downstream deployment scenarios, especially when balancing predictive accuracy against throughput constraints.

\section{Limitations}
\label{sec:limits}
This study isolates two complementary generalization regimes—geometric interpolation/extrapolation and chemical extrapolation—using a curated dataset of ten crystalline materials across four radii. While representative, this selection captures only a limited region of compositional and structural diversity present in real-world materials. All models are evaluated in a zero-shot setting with default decoding configurations, without fine-tuning, retrieval augmentation, or domain adaptation, which may underrepresent their full capabilities.

Evaluation emphasizes first-order structural properties such as lattice constants, density, and stoichiometry, along with a single volumetric consistency index. Higher-order descriptors—including phonon spectra, band topology, or symmetry-preserving deformations—are not considered. The analysis focuses on static prediction quality and does not measure model responsiveness to feedback, learning curves under domain supervision, or variance across decoding seeds. Lastly, benchmark inputs remain synthetic and canonicalized; performance on raw experimental or noisy structural data remains unexplored.

\section{Conclusion}
\label{sec:conc}
This work introduces a new dataset and two complementary benchmarks—SE and CE—that isolate geometric interpolation and chemical extrapolation in crystallographic prediction. The evaluations reveal that current vision–language foundation models struggle to internalize core physical principles, as evidenced by high relative errors, substantial degradation in transfer settings, and disrupted inter-property correlations. The prevalence of hallucinated outputs and violations of basic physical laws further underscores the limitations of purely data-driven training in scientific domains.

To advance reliability and generalization, future models must incorporate explicit physical constraints, symmetry priors, and uncertainty-aware reasoning. The proposed benchmarks provide a reproducible and physically grounded testbed for evaluating model robustness in structured scientific settings. By bridging multimodal language understanding with domain-specific inductive biases, this work aims to foster the development of more trustworthy foundation models for materials science and beyond.

\bibliographystyle{ieeetr}
\bibliography{main}

\newpage
\clearpage
\appendix

\section{Appendix}
\label{sec:appendix}

\subsection{Crystal Parameters}

\label{appen:crystal}
\paragraph{Silver (Ag).}  
Silver adopts an FCC lattice with lattice constant $a = 4.0857$~\AA. The cubic crystal belongs to space group Fm$\overline{3}$m (No.~225), Pearson symbol cF4, and Schoenflies notation $O_h^5$. A single Ag atom occupies the origin of the primitive cell~\citep{silver2002crc}.

\paragraph{Gold (Au).}  
Gold similarly adopts an FCC arrangement with lattice constant $a = 4.0782$~\AA. It crystallizes in space group Fm$\overline{3}$m (No.~225), reflecting equivalent high symmetry. One Au atom resides at the (0,0,0) position within the unit cell~\citep{gold2002crc}.

\paragraph{Methylammonium Lead Iodide (CH$_3$NH$_3$PbI$_3$).}  
The hybrid perovskite CH$_3$NH$_3$PbI$_3$ forms a pseudo-cubic lattice with parameters $a = 6.290$~\AA, $b = 6.274$~\AA, $c = 6.297$~\AA\ and angles close to $90^\circ$. It crystallizes in space group $P1$ (No.~1), accommodating slight distortions and dynamic disorder typical of organic–inorganic frameworks~\citep{walsh2019hybrid}.

\paragraph{Hematite (Fe$_2$O$_3$).}  
Hematite (Fe$_2$O$_3$) exhibits a rhombohedral structure with lattice constants $a = b = 5.0346$~\AA, $c = 13.7473$~\AA, and angles $\alpha = \beta = 90^\circ$, $\gamma = 120^\circ$. It belongs to space group $R\overline{3}c$ (No.~167), underpinning its antiferromagnetic and catalytic properties~\citep{finger1980fe2o3}.

\paragraph{Molybdenum Disulfide (MoS$_2$).}  
Molybdenum disulfide (MoS$_2$) adopts a layered hexagonal lattice with parameters $a = 3.1604$~\AA, $c = 12.295$~\AA, and angles $\alpha = \beta = 90^\circ$, $\gamma = 120^\circ$. It crystallizes in space group $P6_3/mmc$ (No.~194), reflecting its van der Waals–bonded layers~\citep{wyckoff1963mos2, grau2002mos2}.

\paragraph{Galena (PbS).}  
Galena (PbS) forms a rock-salt–type FCC structure with lattice constant $a = 5.9362$~\AA. The cubic crystal belongs to space group Fm$\overline{3}$m (No.~225), with Pb and S atoms occupying alternating FCC sites~\citep{wyckoff1963pbs}.

\paragraph{Cassiterite (SnO$_2$).}  
Cassiterite (SnO$_2$) displays a tetragonal rutile–type lattice with constants $a = 4.738$~\AA, $c = 3.1865$~\AA. It crystallizes in space group $P4_2/mnm$ (No.~136) and features an oxygen sublattice coordinating the Sn atoms~\citep{baur1971sno2}.

\paragraph{Strontium Titanate (SrTiO$_3$).}  
Strontium titanate (SrTiO$_3$) crystallizes in a cubic perovskite structure with lattice constant $a = 3.9053$~\AA\ and space group $Pm\overline{3}m$ (No.~221). Its ideal symmetry underlies its prototypical ferroelectric and quantum paraelectric behavior~\citep{mitchell2000srtio3}.

\paragraph{Titanium Dioxide (TiO$_2$—Anatase).}  
Anatase TiO$_2$ exhibits a body-centered tetragonal structure with $a = 3.7842$~\AA, $c = 9.5146$~\AA. It belongs to space group $I4_1/amd$ (No.~141), characteristic of the anatase polymorph’s photocatalytic activity~\citep{horn1972tio2}.

\paragraph{Zinc Oxide (ZnO—Zincite).}  
Zinc oxide (ZnO) in the zincite phase adopts a hexagonal wurtzite lattice with parameters $a = 3.2495$~\AA, $c = 5.2069$~\AA\ and space group $P6_3mc$ (No.~186). This polar structure underpins its piezoelectric and optoelectronic applications~\citep{wyckoff1963zno}.

\subsection{Additional Metric Definitions}
\label{appen:extendMetric}
\paragraph{Absolute‐error (angles).}
For each primitive‐cell angle 
\(\theta_p\in\{\alpha_p,\beta_p,\gamma_p\}\),
\[
\lvert\Delta\theta_p\rvert = \bigl\lvert\theta_p^{\mathrm{gen}} - \theta_p^{\mathrm{ref}}\bigr\rvert.
\]

\paragraph{Per‐example mean error.}
If an example contains the set of properties \(P\), then
\[
\overline{\%\Delta} = \frac{1}{\lvert P\rvert}\sum_{p\in P}\%\Delta_p.
\]

\paragraph{Format faithfulness.}
Let \(\mathcal F_{\mathrm{ref}}\) and \(\mathcal F_{\mathrm{gen}}\) be the
non-null field sets, and \(\mathcal F_{\cap} = \mathcal F_{\mathrm{ref}}\cap\mathcal F_{\mathrm{gen}}\).  The following definitions are considered:
\[
\begin{aligned}
S_{\mathrm{presence}} &= \frac{\bigl|\mathcal F_{\cap}\bigr|}{\bigl|\mathcal F_{\mathrm{ref}}\bigr|}\\[6pt]
S_{\mathrm{type}} &= \frac{1}{\bigl|\mathcal F_{\cap}\bigr|}
  \sum_{f\in\mathcal F_{\cap}}
    \mathbf{1}\!\bigl(\mathrm{type}_{\mathrm{gen}}(f)=\mathrm{type}_{\mathrm{ref}}(f)\bigr) ,
\end{aligned}
\]
and
\[
S_{\mathrm{format}} = 0.7\,S_{\mathrm{presence}} + 0.3\,S_{\mathrm{type}}.
\]

\end{document}